\documentclass{article}

\usepackage[nonatbib, preprint]{neurips_2021}

\makeatletter
\newcommand{\printfnsymbol}[1]{%
  \textsuperscript{\@fnsymbol{#1}}%
}
\makeatother

\usepackage[utf8]{inputenc} 
\usepackage[T1]{fontenc}    
\usepackage{hyperref}       
\usepackage{url}            
\usepackage{booktabs}       
\usepackage{amsfonts}       
\usepackage{nicefrac}       
\usepackage{microtype}      
\usepackage{xcolor}         
\usepackage{mathrsfs}
\usepackage{graphicx}     
\usepackage{amsmath}
\usepackage{caption}
\usepackage{subcaption}
\usepackage{amssymb}

\title{ERNIE-Tiny : A Progressive Distillation Framework for Pretrained Transformer Compression}

\author{
Weiyue Su\thanks{Equal Contribution.}, Xuyi Chen\printfnsymbol{1}, Shikun Feng, Jiaxiang Liu, Weixin Liu, \\
\textbf{Yu Sun, Hao Tian, Hua Wu, Haifeng Wang} \\ 
  Baidu Inc., China \\
  \tt \{suweiyue, chenxuyi, fengshikun01, liujiaxiang, liuweixin \\
  \tt sunyu02, tianhao, wu\_hua, wanghaifeng\}@baidu.com }

\begin{document}

\maketitle

\begin{abstract}
Pretrained language models (PLMs) such as BERT \cite{devlin2018bert} adopt a training paradigm which first pretrain the model in general data and then finetune the model on task-specific data, and have recently achieved great success. However, PLMs are notorious for their enormous parameters and hard to be deployed on real-life applications. Knowledge distillation \cite{hinton2015distilling} has been prevailing to address this problem by transferring knowledge from a large teacher to a much smaller student over a set of data. We argue that the selection of thee three key components, namely teacher, training data, and learning objective, is crucial to the effectiveness of distillation. We, therefore, propose a four-stage progressive distillation framework ERNIE-Tiny to compress PLM, which varies the three components gradually from general level to task-specific level. Specifically, the first stage, \textbf{General Distillation}, performs distillation with guidance from pretrained teacher, gerenal data and latent distillation loss. Then, \textbf{General-Enhanced Distillation} changes teacher model from pretrained teacher to finetuned teacher. After that, \textbf{Task-Adaptive Distillation} shifts training data from general data to task-specific data. In the end, \textbf{Task-Specific Distillation}, adds two additional losses, namely Soft-Label and Hard-Label loss onto the last stage. Empirical results demonstrate the effectiveness of our framework and generalization gain brought by ERNIE-Tiny. In particular, experiments show that a 4-layer ERNIE-Tiny maintains over 98.0\% performance of its 12-layer teacher BERT$_{base}$ on GLUE benchmark, surpassing state-of-the-art (SOTA) by 1.0\% GLUE score with the same amount of parameters. Moreover, ERNIE-Tiny achieves a new compression SOTA on five Chinese NLP tasks, outperforming BERT$_{base}$ by 0.4\% accuracy with 7.5x fewer parameters and 9.4x faster inference speed. 
\end{abstract}

		\if 
		KD tries to leverage the knowledge learnt by teacher model to improve student's performance by mimicking intermediate layer or final logits of teacher and student.
		Distillation has become a common practice in compressing large-scale Transformer based language models. Some previous works focus on task-agnostic distillation, while other approaches perform distillation in both pretraining and finetuning stages. However, these two stages share very few common points among the three key distillation factors, which are training data, teacher model, and distillation objective. The huge difference between the consecutive distillation stages hurts the performance. 
		Inspired by the idea of xxx, we design ERNIE-Tiny, a multi-stage distillation framework to bridge the gap and smooth the learning curve by upgrading distillation process from general to specific data step by step.
		One key stage in the proposed framework is using pretraining corpus as transfer data in task-specific distillation. We find out a finetuned teacher can extract more effective task-specific knowledge from massive unsupervised pretraining dataset than a pretrained teacher, and the student model achieves better performance on downstream tasks. 
		Experiments show that our tiny model with 4 layers is empirically effective and maintains over 98.0\% performance of its teacher BERT$_{base}$ on GLUE benchmark. It also surpasses the state-of-the-art baseline model by 1.0\% GLUE score with the same parameters.
		Specially, if we change the teacher to ERNIE2.0$_{base}$, a better language model than BERT$_{base}$, ERNIE-Tiny achieves a new compression SOTA on five Chinese NLP tasks, exceeds TinyBERT by 1.4\% average score, and even outperforms BERT$_{base}$ by 0.4\%, while our model has 7.5x smaller parameters and 9.4x faster inference speed.
		\fi

\vspace{2mm}
\section{Introduction}
    Transformer-based pretrained language models (PLMs) \cite{devlin2018bert, roberta, albert, sun2019ernie, bart, xlm, xlm-r} have brought significant improvements to the field of Natural Language Processing (NLP). Their training process that first pretrains model on general data and then finetunes on task-specific data has set up a new training paradigm for NLP. However, the performance gains come with the massive growth in model sizes \cite{gpt3, t5, fedus2021switch, megatron-lm} which causes high inference time and storage cost. It becomes the main obstacle for industrial application, especially for deploying on edge devices. 
\begin{figure}[htbp]
\centering
\begin{minipage}{.5\textwidth}
  \centering
      \includegraphics[width=0.9\textwidth]{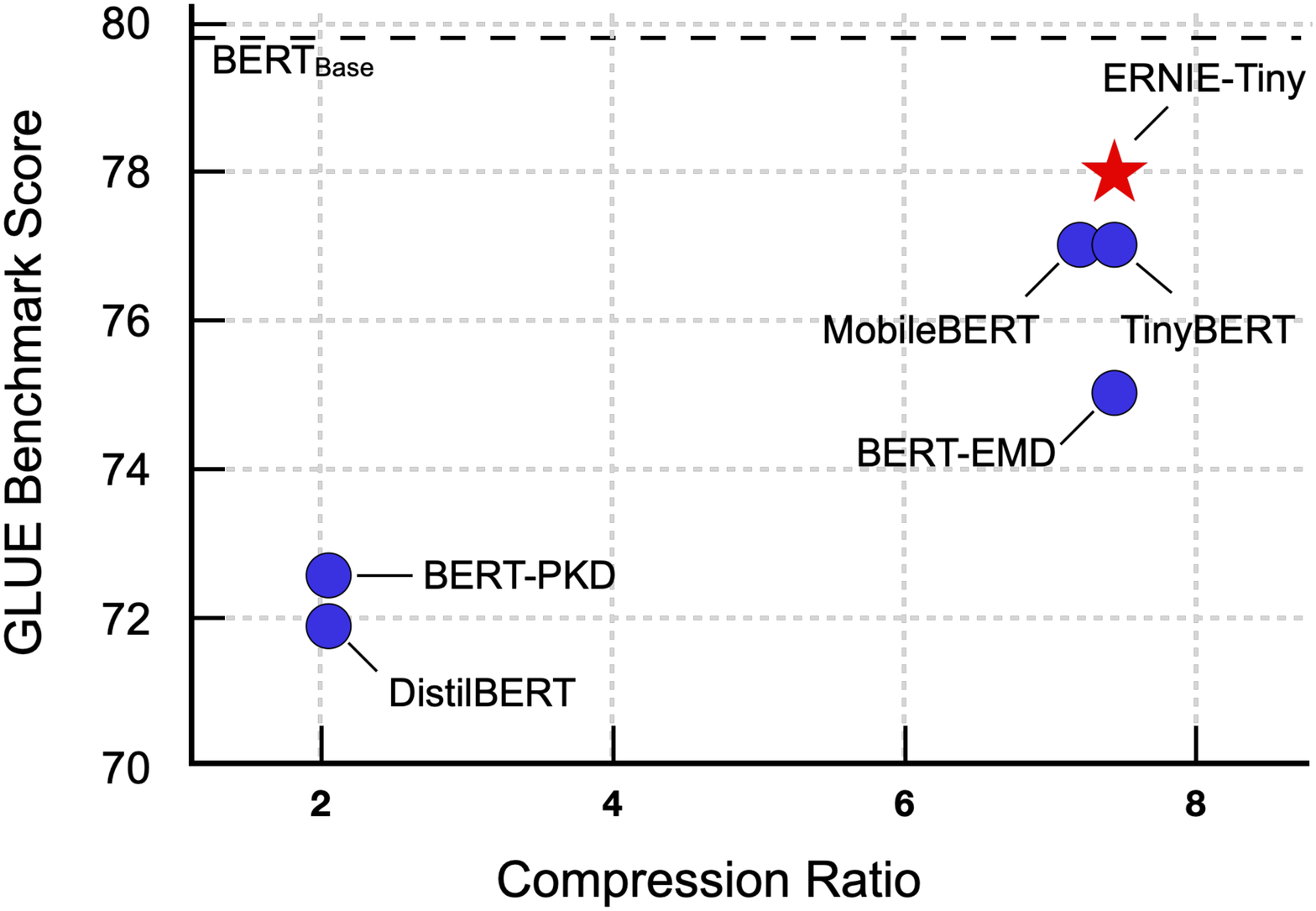}
  \captionsetup{width=0.95\textwidth}
  \setlength{\abovecaptionskip}{0.1mm}
  \caption{GLUE score of different distillation methods. Performance of the teacher, BERT$_{base}$, is shown in dash line.}  
  \setlength{\belowcaptionskip}{-5cm}
  \label{fig:tiny-a}
\end{minipage}%
\begin{minipage}{.5\textwidth}
  \centering
  \includegraphics[width=0.9\textwidth]{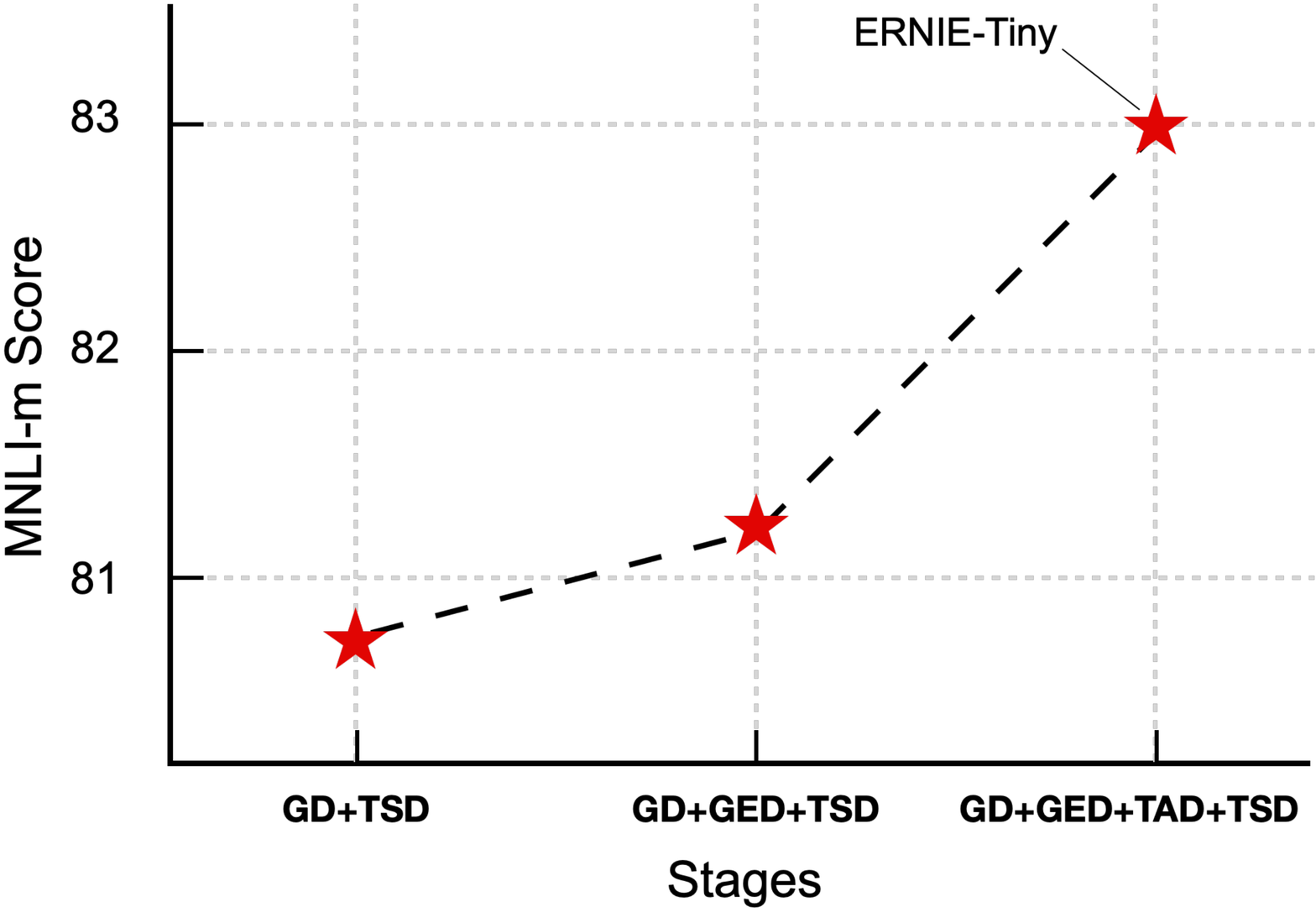}
  \captionsetup{width=0.95\textwidth}
  	\setlength{\abovecaptionskip}{-0.1mm}
    \caption{Contribution of general-enhanced distillation and task-adaptive distillation. MNLI-m results are reported on dev set.}
  \setlength{\belowcaptionskip}{-5cm}
\label{fig:tiny-b}
\end{minipage}
\vspace{-9mm}
\end{figure}

	There are some recent efforts such as Knowledge Distillation (KD) \cite{hinton2015distilling, urban2016deep, ba2013deep}, quantization\cite{kim2019qkd, shin1909empirical, wei2018quantization}, and weights pruning\cite{wang2018packing, han2015learning, sindhwani2015structured} trying to tackle this problem. 
	KD, in particular, aims to transfer knowledge from one network called teacher model to another called student model by training student under the guidance of teacher. Typically, teacher is a model with more parameters and capable of achieving high accuracy, whereas student is a model with significantly fewer parameters and requires much less computation. Once trained, the student model maintains teacher's performance while massively reducing inference time and storage demand, and can be deployed in real-life applications. KD can be applied on either or both of pretrain and finetune stages. For example, MiniLM \cite{minilm} and MobileBert \cite{mobilebert} apply KD on pretrain stage while \cite{pkd} applies KD on finetune stage. Moreover, TinyBERT \cite{tinybert} and DistilBert \cite{distill-bert} perform KD on both pretrain and finetune stages. In particular, they employ pretrained teacher to provide guidance during pretrain stage and choose finetuned teacher during finetune stage, where pretrained teacher is the teacher model trained on general data and finetuned teacher is obtained by finetuning pretrained teacher on task-specific data. 

\begin{figure*}[htbp]
	\centering
		\includegraphics[width=1.0\textwidth]{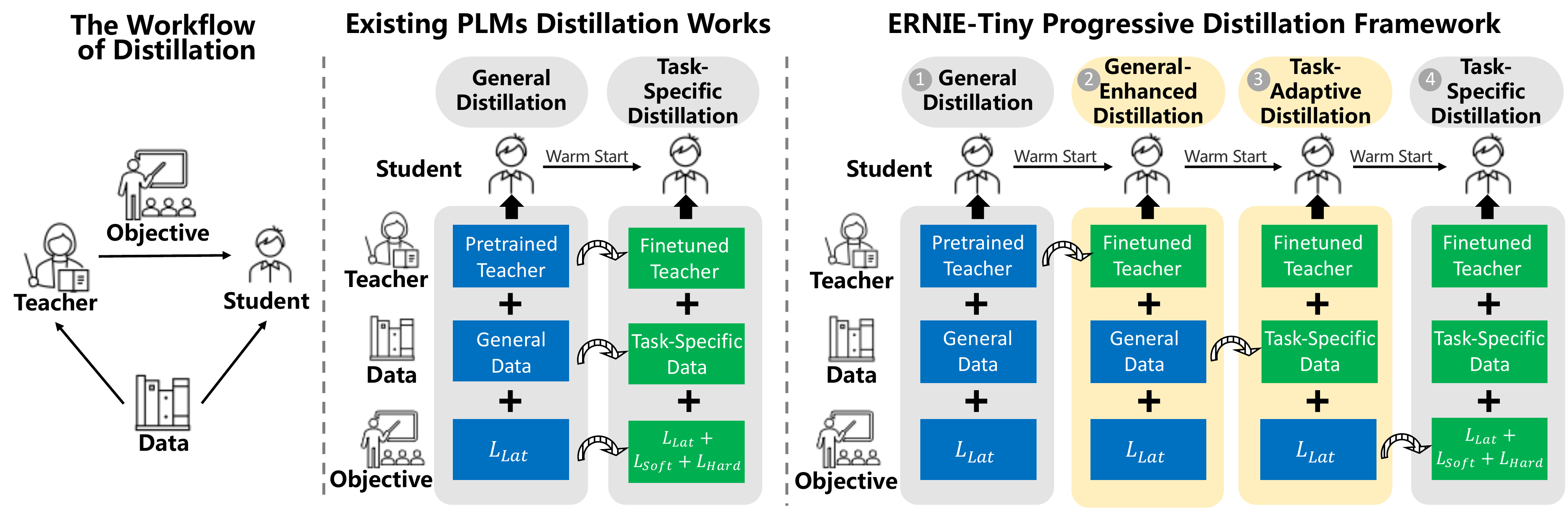}
	\caption{The comparison between existing works and ERNIE-Tiny. The curly shaded arrow indicates the change of the three key components (i.e. Teacher, Data, and Objective). \textbf{Left: Workflow of Distillation.} Teacher transfers its knowledge to student through data and objective.
	\textbf{Middle: Workflow of Existing Works.} All of the three components shift between the two stages. \textbf{Right: Workflow of ERNIE-Tiny}. ERNIE-Tiny carefully designs the distillation framework such that only one component is changed between any two consecutive stages.}
	\label{fig:Distillation}
	\vspace{-4mm}
\end{figure*}

	However, existing works suffer from pretrain-finetune distillation discrepancy consisting of the difference of training data, teacher model, and learning objective between pretrain phase and finetune phase.
	Specifically, training data is shifted from general data to task-specific data, teacher is changed from pretrained teacher to finetuned teacher, and learning objective is altered differently according to their own decisions. 
	We argue that this transition hurts the effectiveness of distillation.
	We, therefore, propose a four-stage progressive distillation framework ERNIE-Tiny to alleviate this problem and our method outperforms several baselines as shown in Figure~\ref{fig:tiny-a}. ERNIE-Tiny attempts to smooth this pretrain-finetune transition by gradually altering teacher, learning objective, and training data from general level to task-specific level.
	Akin to pretrain distillation at existing works, \textbf{General distillation (GD)} performs distillation with pretrained teacher on general data. Following previous works \cite{tinybert, pkd, mobilebert, distill-bert}, we utilize latent distillation ($\mathcal{L}_{Lat}$) as our learning objective. Then, by altering teacher from pretrained teacher to finetuned teacher, ERNIE-Tiny introduces \textbf{General-Enhanced Distillation (GED)} which distills with finetuned teacher and $\mathcal{L}_{Lat}$ on general data. After that, through changing training data from general data to task-specific data, ERNIE-Tiny presents \textbf{Task-Adaptive Distillation (TAD)} which distills with finetuned teacher and $\mathcal{L}_{Lat}$ on task-specific data. Finally, ERNIE-Tiny concludes the training process with \textbf{Task-Specific Distillation (TSD)} through adding new learning objectives, namely Soft-Label Distillation ($\mathcal{L}_{Soft}$) and Hard-Label loss ($\mathcal{L}_{Hard}$) which represents the task-specific finetune loss such as cross-entropy for classification downstream task. Note that TSD is similar to the finetune distillation at existing works. Figure~\ref{fig:Distillation} compares the workflow of existing works and ERNIE-Tiny.

    
    
	Notably, general-enhanced distillation provides finetuned teacher guidance not on task-specific data as what existing works do, but on general data. Compared with existing works, general-enhanced distillation allows student to absorb task-specific knowledge through general data, improving the effectiveness of distillation and generalization of student model \cite{laine2016temporal, sajjadi2016regularization, miyato2018virtual, goodfellow2014explaining}. Empirical results show that with general-enhanced distillation, ERNIE-Tiny outperforms the baseline on out-of-domain datasets, demonstrating the generalization gain brought by general-enhanced distillation. In addition, general data can be regarded as additional data to task-specific data. We conduct experiments to show that the effect of general-enhanced distillation is more significant on low-resource tasks. Moreover, task-adaptive distillation is introduced between general-enhanced distillation and task-specific distillation, serving as a bridge to smooth the transition between those two stages. We conduct experiments to show the performance gain brought by this stage. Figure~\ref{fig:tiny-b} demonstrates the contribution of general-enhanced distillation and task-specific distillation by gradually adding general-enhanced distillation and task-specific distillation.
	
    The \textbf{main contributions} of this work are as follows: \textbf{1)} We propose a four-stage progressive learning framework for language model compression called ERNIE-Tiny to smooth the distillation process by gradually altering teacher, training data, and learning objective. \textbf{2)} To our knowledge, leveraging finetuned teacher with general data is the first time introduced in PLM distillation, helping student capture task-specific knowledge from finetuned teacher and improving generalization of student. \textbf{3)} A 4-layer ERNIE-Tiny keeps over 98.0\% performance of its 12-layer teacher BERT$_{base}$ on GLUE benchmark and exceeds state-of-the-art (SOTA) by 1.0\% GLUE score. In Chinese datasets, 4-layer ERNIE-Tiny, harnessed with a better teacher, outperforms BERT$_{base}$ by 0.4\% accuracy with 7.5x fewer parameters and 9.4x faster inference speed.
    \vspace{-3mm}
	
	\if
	Since the release of BERT \cite{devlin2018bert}, large-scale Transformer model has become a fundamental building block for Natural Language Processing (NLP) tasks. The rapid growth in model size becomes the main barrier for industrial application.
	Model compression and acceleration have been widely studied by the Machine Learning community, especially in Computer Vision (CV). 
	The residual structure of deep Transformers bridge the gap between the natural language processing model and convolution-based CV model, making it possible for the application of the ResNet-based acceleration method on NLP tasks. 
	Among many of the acceleration and compression methods, knowledge distillation \cite{hinton2015distilling} is the most widely adopted.
	In knowledge distillation, a compact student model is forced to mimic the predicted probability of teacher-model. Also recent work \cite{fitnets} shows that with intermediate hidden mimicing, student model can achieve better performance.
	
	There are two main approaches for model distillation: task-agnostic and task-specific distillation.
	Most of the recent works on Transformer distillation focus on a task-agnostic way of model compression. 
	Google \cite{well-read} shows that using a task-agnostic pretrained model for student model initialization achieves better distillation performance. 
	Huggingface \cite{distill-bert} trains a general proposed compact model by joint training mask language model (MLM) loss and knowledge distillation loss. 
	In terms of industrial application, however, we might encounter hundreds of different tasks and each has its requirement. 
	For example, online query intention detection is relatively easy but requires low latency, Question Answering (QA) task for the long paragraph is much more complicated, but can be processed in an off-line fashion. Semantic retrieval of a large amount of document requires high throughput. 
	To achieve maximum gains on each of these tasks, task-specific distillation is a more viable solution than the one-size-fit-all approach. 
	Some other researches \cite{chen2020adabert} focus on task-specific distillation, among which LSTM/CNN is often chosen as student architecture.
	It is intuitive to choose the right model for each task, but in the recent development of NLP models, we can see a unifying tendency that Transformer \cite{vaswani2017attention}  becomes a general-purpose model architecture dominating leaderboard from NLU \cite{glue} to text-generation. 
	Besides, some researchers focus on Transformer latency optimization, utilizing the parallel nature of Transformer to achieve extreme acceleration and compression. These facts suggest that as long as task-specific knowledge is given, Transformer model could efficiently be applied to many of the downstream tasks.
	As for the general web-crawled pretraining corpus, it is always used in an unsupervised manner, the potential of task-specific usage of these data has to be discovered. 
	
	As the literature \cite{bengio2009curriculum} says, humans and animals learn much better when the courses are organized in a meaningful order which goes from easy to hard. Figure \ref{fig:motivation} illustrates the different learning curves between traditional distillation (blue line) and our curriculum learning-based distillation (red line). When the learning process meets the difficult knowledge zone which is rendered as the yellow rectangle in Figure \ref{fig:motivation}, the traditional distillation method always fails to achieve high performance due to setting only a hard course in such a difficult zone. The proposed approach is to follow the setup of curriculum learning \cite{bengio2009curriculum}. We carefully set up a certain number of right courses to help the student model to cross the difficult zone, and achieve better performance which is more close to the teacher model.
	
	In this work, we explore an effective approach for task-specific distillation. 
	In addition to pretrain a general proposed compact model on an unsupervised corpus, we train teacher for each task to perform distillation on the pretraining corpus. 
	Some tasks require models to have sentence pairs as inputs, such as NLI and QA. To build the input of the teacher model with such requirements, we concatenate adjacent sentences from pretraining corpus. 
	Also, we applied intermediate distillation and attention distillation during the whole process. 
	In our distillation setup, the teacher model act as a feature extractor. The general data is used twice, the first time in the task-agnostic pretraining stage, the second time in the task-specific distillation stage. 
	We provide empirical studies to show that task-specific teacher can extract more useful knowledge from the unsupervised corpus and improve student model performance.
	To explore the performance gain limit of task-specified distillation, we also set up a self-distillation scenario where the student has the same model size as the teacher model. we find that self-training with a task-specific teacher helps the student model to achieve better performance than the teacher.
	
	In the following sections, we give an overview of the development of language model (LM) compression and semi-supervised learning in section 2 and introduction of the proposed distillation framework in section 3. Section 4-6 include the detailed experiment setup and empirical study results showing effectiveness of the proposed method. Our main contributions are:
	
	\begin{enumerate}
		\item We propose a four-stage curriculum learning framework for language model compression, which we call ERNIE-Tiny. The detailed analysis are given on several key designs: choice of the teacher model, multi-stage distillation, intermediate distillation and student capacity.
		
		\item To our knowledge, task-specific distillation leveraging pretraining corpus is the first time introduced in pretrained language model distillation, which helps to achieve better compression performance. 
		
		\item Our curriculum learning distillation method surpasses the previous language model distillation methods,  and achieves new compression SOTA on multiple types of NLP tasks in both English and Chinese.
	\end{enumerate}
	\fi


\section{Related Work}

\paragraph{Pretrained Language Models} Pretrained language models are learned on large amounts of text data and then finetuned to adapt to specific tasks. BERT \cite{devlin2018bert} proposes to pretrain a deep bidirectional Transformer. RoBERTa \cite{roberta} achieves strong performance by training longer steps using large batch size and more text data. ERNIE \cite{sun2019ernie} \cite{sun2019ernie20} proposes to pretrain the language model on an enhanced mask whole word objective and further employs continue learning strategy. XL-Net \cite{xlnet} introduces permutation language modeling objective to alleviate the pretrain-finetune discrepancy which is caused by artificial symbols like [MASK] used during pretraining but is absent from real data at finetuning. SpanBERT \cite{spanbert} improves BERT by incorporating span information. Besides, \cite{xlm} \cite{xlm-r} leverages parallel bilingual data to train the multi-lingual language model. Recent works (\cite{megatron-lm}, \cite{gpt3}, \cite{kaplan2020scaling}) observe the trend that increasing model size also leads to lower perplexity. \cite{megatron-lm} proposes to train a billion level parameters model for language understanding using a carefully designed distributed computation system. Switch-transformer \cite{fedus2021switch} simplifies and improves over Mixture of Experts \cite{shazeer2017outrageously} and trains a trillion parameters language model. However, \cite{kovaleva2019revealing} shows the parameters are redundant in those models and the performance can be kept even when the computational overhead and model storage is reduced. Moreover, the training cost of those models also raises serious environmental concerns \cite{strubell2019energy}. 

\paragraph{Knowledge Distillation}
Knowledge distillation \cite{hinton2015distilling, minilm} aims to train a small student model with soft labels and intermediate representations provided by the large teacher model. Concurrently to our work, \cite{bert-emd} utilizes Earth Mover’s Distance \cite{rubner2000earth} to formulate the distance between the teacher and student networks.\cite{tinybert} proposes TinyBERT on the general distillation and task-specific distillation stages.  \cite{distill-bert} proposes DistilBERT, which successfully halves the depth of BERT model by knowledge distillation in the pretrain stage and an optional finetune stage. \cite{pkd} distills BERT into a shallower student through knowledge distillation only in the finetune stage. \cite{minilm} proposes to compress teacher by mimicking self-attention and value relation in the pretrain stage.  \cite{mobilebert} utilizes bottleneck structures to distill self-attentions and feed-forward networks. In contrast to these existing literature, we argue that the pretrain-finetune distillation discrepancy exists. Specifically, the pretrain-finetune distillation discrepancy is caused by training data shift, teacher model alteration and learning objective change. Therefore, we propose a progressive distillation framework ERNIE-Tiny to compress PLM. Through this progressive distillation framework, the discrepancy of distillation can be alleviated and the performance of the distilled student can be improved. Table~\ref{tab:ernie-tiny learning} summarizes the differences between our framework and previous works. 
\vspace{-3mm}

\begin{table}[h]
	\centering
	 \scalebox{0.7}{
	\begin{tabular}{ cccccccccc }
		\toprule
		Stage & Teacher & Data & ERNIE-Tiny & BERT-EMD & TinyBERT & DistillBert & BERT-PKD & MiniLM & MobileBert \\
		\midrule
		GD & pretrained & General  & $\mathcal{L}_{Lat}$ & $\mathcal{L}_{Lat}$  & $\mathcal{L}_{Lat}$   & $\mathcal{L}_{Lat}$+$\mathcal{L}_{Soft}$ & -  & $\mathcal{L}_{Lat}$ & $\mathcal{L}_{Lat}$+$\mathcal{L}_{Soft}$ \\
		GED & finetuned & General&  $\mathcal{L}_{Lat}$ &- & - & -  & - & - & - \\
		TAD & finetuned & Task-Specific & $\mathcal{L}_{Lat}$ &- & - & -  & - & - & -  \\
		TSD & finetuned & Task-Specific  & $\mathcal{L}_{L+S+H}$  & $\mathcal{L}_{L+S+H}$ & $\mathcal{L}_{L+S+H}$ & $\mathcal{L}_{L+S+H}$& $\mathcal{L}_{L+S+H}$ & $\mathcal{L}_{Hard}$ & $\mathcal{L}_{Hard}$ \\
		\bottomrule
	\end{tabular}
	}
	\vspace{1mm}
	\caption{Comparison with previous PLM distillation approaches. Latent Distillation ($\mathcal{L}_{Lat}$) represents distillation loss on the attributes at intermediate layers and it varies on different methods (e.g hidden states and attention distribution in TinyBERT and BERT-EMD; attention distribution and attention value relation in MiniLM). Soft-Label Distillation ($\mathcal{L}_{Soft}$) denotes distillation on soft target probabilities from the teacher model. As all methods adopt Hard-Label loss ($\mathcal{L}_{Hard}$) in TSD, for simplicity, we denote  $\mathcal{L}_{L+S+H}=\mathcal{L}_{Lat}+\mathcal{L}_{Soft}+\mathcal{L}_{Hard}$.}
    \label{tab:ernie-tiny learning}
\end{table}
\vspace{-7mm}

\if{}
\begin{table}[h]
	\centering
	 \scalebox{0.7}{
	\begin{tabular}{ cccccccccc }
		\toprule
		Stage & Teacher & Data & ERNIE-Tiny & BERT-EMD & TinyBERT & DistillBert & BERT-PKD & MiniLM & MobileBert \\
		\midrule
		GD & pretrained & General  & $\mathcal{L}_{Lat}$ & $\mathcal{L}_{Lat}$  & $\mathcal{L}_{Lat}$   & $\mathcal{L}_{Lat}$+$\mathcal{L}_{Soft}$ & -  & $\mathcal{L}_{Lat}$ & $\mathcal{L}_{Lat}$+$\mathcal{L}_{Soft}$ \\
		GED & finetuned & General&  $\mathcal{L}_{Lat}$ &- & - & -  & - & - & - \\
		TAD & finetuned & Task-Specific & $\mathcal{L}_{Lat}$ &- & - & -  & - & - & -  \\
		TSD & finetuned & Task-Specific  & $\mathcal{L}_{Lat}$+$\mathcal{L}_{Soft}$  & $\mathcal{L}_{Lat}$+$\mathcal{L}_{Soft}$ & $\mathcal{L}_{Lat}$+$\mathcal{L}_{Soft}$ & $\mathcal{L}_{Lat}$+$\mathcal{L}_{Soft}$& $\mathcal{L}_{Lat}$+$\mathcal{L}_{Soft}$ & \checkmark & \checkmark \\
		\bottomrule
	\end{tabular}
	}
	\vspace{1mm}
	\caption{Comparison with previous Transformer based PLM distillation approaches. Latent Distillation ($\mathcal{L}_{Lat}$) represents distillation loss on the attributes at intermediate layers and it varies on different methods (e.g hidden states and attention distribution in TinyBERT and BERT-EMD; attention distribution and attention value relation in MiniLM). Soft-Label Distillation ($\mathcal{L}_{Soft}$) denotes distillation on soft target probabilities from the teacher model. As all methods adopt Hard-Label loss in TSD, we do not show them for simplicity.}
    \label{tab:ernie-tiny learning}
\end{table}
\vspace{-2mm}
\fi
\vspace{-3mm}
\section{Proposed Framework}
Distillation aims to use the pretrained teacher $T$ to teach a student model $S$ that is usually much smaller as shown in the left part of Figure~\ref{fig:Distillation}. In our setting, besides the labeled task-specific data $D_t$, we also have large-scale unlabeled data which we call general data $D_g$ from which the teacher is pretrained. In practice, those unlabeled data can be easily obtained from online webs such as C4, Wikipedia and BookCorpus \cite{zhu2015aligning} used in pretrained language model (\cite{t5}, \cite{devlin2018bert}). To combine those data and teacher knowledge smoothly, we devise a progressive distillation framework, containing general distillation, general-enhanced distillation, task-adaptive distillation and task-specific distillation sequentially. Those four stages vary the three key distillation components, namely training data, teacher model and learning objective gradually from general level to task-specific level as shown in Figure~\ref{fig:Distillation}. To better explain those methods, we first show the background and discuss the distillation framework in detail.




\subsection{Background: Transformer Backbone}

The Transformer architecture \cite{vaswani2017attention} is a highly modularized neural network, where each Transformer layer consists of two sub-modules, namely the multi-head self-attention ($\mathrm{MHA}$) and position-wise feed-forward network ($\mathrm{FFN}$).  Transformer encodes contextual information for input tokens. The input embeddings $\{\mathbf{x}\}^{s}_{i=1}$ for sample $x$ are packed together into
$\mathbf{H}_{0}=\left[\mathbf{x}_{1}, \cdots, \mathbf{x}_{s}\right ] $ , where $s$ denotes the input sequence length. Then stacked Transformer blocks iteratively compute  the encoding vectors as $\mathbf{H}_{l}=\mathrm { Transformer }_{l}\left(\mathbf{H}_{l-1}\right), l \in[1, L]$, and the Transformer is computed as:
\vspace{-1mm}
\begin{equation*}
\scalebox{0.9}{
\begin{math}
\begin{aligned}
\mathbf{A}_{l, a}&=\operatorname{MHA}_{l, a}(
\mathbf{H}_{l-1} \mathbf{W}_{l, a}^{Q}, \mathbf{H}_{l-1} \mathbf{W}_{l, a}^{K}), \\
\mathbf{H}_{l-1}' & = \text {LayerNorm}(\mathbf{H}_{l-1}+ (\mathop{\|}_{a=1}^h \mathbf{A}_{l, a}  (\mathbf{H}_{l-1} \mathbf{W}_{l, a}^{V})) \mathbf{W}_{l}^{O}) , \\
\mathbf{H}_{l} & = \text {LayerNorm}\left(\mathbf{H}_{l-1}'+\mathrm{FFN}\left(\mathbf{H}_{l-1}'\right)\right),
\end{aligned}
\end{math}}
\end{equation*}

\vspace{-3mm}


where the previous layer’s output $\mathbf{H}_{l-1} \in \mathbb{R}^{ s \times d}$ is linearly projected to a triple of queries, keys and values using parameter matrices $\mathbf{W}^{Q}_{l,a}, \mathbf{W}^K_{l,a}, \mathbf{W}^{V}_{l,a} \in \mathbb{R}^{d \times d'}$, where $d$ denotes the hidden size of $\mathbf{H}_l$ and $d'$ denotes the hidden size of each head's dimension. $\mathbf{A}_{l,a} \in \mathbb{R}^{s \times s}$  indicates the attention distributions for the $a$-th head in layer $l$, which is computed by the scaled dot-product of queries and keys respectively. $h$ represents the number of self-attention heads. $ \| $ denotes concatenate operator along the head dimension. $\mathbf{W}_l^{O} \in \mathbb{R}^{d\times d}$ denotes the linear transformer for the output of the attention module. $\mathrm{FFN}$ is composed of two linear transformation function including mapping the hidden size of $\mathbf{H}_{l-1}'$ to $d_{ff}$ and then mapping it back to ${d}$.

\subsection{General Distillation and General-Enhanced Distillation} 

\paragraph{General Distillation}As shown in Figure \ref{fig:Distillation}, ERNIE-Tiny employs general distillation and general-enhanced distillation sequentially. In the general distillation stage, the pretrained teacher helps the student learn knowledge on the massive unlabeled general data with the intermediate representation. The loss is computed as follows:
\begin{equation}
\scalebox{0.9}{
\begin{math}
\begin{aligned}
\label{equ:gd}
    \mathcal{L}_{GD}&=
	\mathop{\mathbb{E}}_{x \sim {D_g}} 
	\underbrace{
	\sum_{l=1}^{L_S} \sum_{a=1}^{h} F( \mathbf{A}_{k,a}^{T_g}(x),  \mathbf{M}_{l,a}^{\quad} \mathbf{A}_{l}^{S}(x)) + \sum_{l=1}^{{L}_{S}} F(  \mathbf{H}_{k}^{T_g}(x), \mathbf{H}_{l}^{S}(x)\mathbf{N}_{l} ) }_{\mathcal{L}_{Lat}^{T_g}(x)}, k=l\times c,\\
\end{aligned}
\end{math}
}
\end{equation}
where $\mathcal{L}_{GD}$ denotes the loss for general distillation on the general data $D_g$ . $L_S$ denotes the number of layers of student model. Considering the number of layers of pretrained teacher ${L}_{T}$ and student model ${L}_{S}$ may not be the same, we set student layers to mimic the representation of every ${c}$ layers of pretrained teacher model, where ${c}={L}_{T}  \mathbin{/}  {L}_{S}$. 
We introduce a mapping matrix $\mathbf{M}_{l} \in \mathbb{R}^{h \times h'}$ to align the number of attention heads for teacher and student's attention heads, $h$ and $h'$, when they do not match.
Similarly, a linear transformation $\mathbf{N}_{l} \in \mathbb{R}^{d \times d'}$ is used when the hidden size $d$ and $d'$ of $\mathbf{H}_{l}^{T} \in \mathbb{R}^{s \times d} $ and $\mathbf{H}_{l}^{S} \in \mathbb{R}^{s \times d'} $ does not match. A metric function $F$ is utilized to measure the distance between teacher and student's representation and guide the distillation process. We choose mean square error as $F$ for our experiment. Put it together, we call the right hand side of Eq.~\ref{equ:gd} latent distillation and denotes it as $\mathcal{L}_{Lat}^{T_g}$ where $T_{g}$ indicates the pretrained teacher (i.e. the guidance $\mathbf{A}^{T_g}$ and $\mathbf{H}^{T_g}$ come from pretrained teacher).

\paragraph{General-Enhanced Distillation} To further exploit the general data, we propose to use the finetuned teacher as a surrogate for task-specific knowledge and perform distillation over general data. And the training loss of general-enhanced distillation is defined as follows:
\begin{equation}
\scalebox{0.9}{
\begin{math}
\begin{aligned}
\label{equ:ged}
	\mathcal{L}_{GED}&=
	\mathop{\mathbb{E}}_{x \sim {D_g}} \mathcal{L}_{Lat}^{T_f}(x) ,
\end{aligned}
\end{math}
}
\end{equation}
where $\mathcal{L}^{T_f}_{Lat}$ indicates that the guidance involved in latent distillation loss comes from finetuned teacher. During general-enhanced distillation, the student is optimized by minimizing the $\mathcal{L}_{GED}$ on general data. 

One benefit of this stage is that the distillation process becomes much smoother. Comparing Eq.~(\ref{equ:ged}) with Eq.~(\ref{equ:gd}), the only change between general distillation and general-enhanced distillation is that we only replace the teacher $T_g$ with $T_f$ among the three components (i.e. teacher, training data, learning objective) while existing works change all of them together at the same time as shown in Figure~\ref{fig:Distillation}. 


Another benefit is that introducing finetuned teacher on general data improves the generalization of student model. 
As the number of task-specific samples is usually much smaller than general data,
having the finetuned teacher generating hidden representations on general data can be used to compensate for the task-specific data sparsity. Those hidden representations extracted from $D_g$ can be regarded as feature augmentation. Although there may be no task-related label information on $D_g$, the hidden representation from finetuned teacher still contains task-specific information. Several works \cite{laine2016temporal, sajjadi2016regularization, miyato2018virtual, goodfellow2014explaining} succeed in using the random image augmentation to improve generalization performance for semi-supervised tasks. The empirical results on generalization gains led by general-enhanced distillation are shown in Section~\ref{sec:ablation}.

\subsection{Task-Adaptive Distillation and Task-Specific Distillation} 
\paragraph{Task-Adaptive Distillation} Task-adaptive distillation is introduced after general-enhanced distillation to start distillation on task-specific data. The task-adaptive distillation loss is devised as following: 
\begin{equation}
\begin{aligned}
\label{equ:tad}
\mathcal{L}_{TAD} = \mathop{\mathbb{E}}_{x \sim {D_t}} \mathcal{L}_{Lat}^{T_f}(x),
\end{aligned}
\end{equation}
where $D_t$ is the task-specific data. Student model is trained by minimizing $\mathcal{L}_{TAD}$. Comparing Eq.(\ref{equ:tad}) with Eq.(\ref{equ:ged}), we see that the difference between general-enhanced distillation and task-adaptive distillation is that the training data is changed from general data to task-specific data.
 
The advantage of proposing the task-specific stage is two-fold. First, continuing with the philosophy of progressive distillation and pretrain-then-finetune paradigm, only the dataset is changed in this stage to smoothen the distillation. Second, as recent work \cite{t5} shows that unsupervised learning on the task-specific data before applying the supervised signal leads to improvement on downstream performance, distillation of hidden representations on task-specific data paves the way for the upcoming task-specific objective learning.
 

\paragraph{Task-Specific Distillation} Task-specific distillation is presented to finish the whole distillation process. Compared with the last stage, this stage includes soft-label and hard-label learning objectives. Soft-label is the logit of teacher and hard-label is the ground-truth label from the task-specific data. Specifically, the loss is computed as follows: 
\vspace{-2mm}
\begin{equation}
\begin{aligned}
\label{equ:td}
\mathcal{L}_{TSD} &= \mathop{\mathbb{E}}_{(x, y) \sim {D_t}} \mathcal{L}_{Lat}^{T_f}(x) + \mathcal{L}^{T_{f}}_{Soft}(x) + \mathcal{L}_{Hard}(x,y), \\
\mathcal{L}_{Soft}^{T_f}(x)&=F_1(z^{T_f}(x), z^S(x)), \\
\mathcal{L}_{Hard}(x, y)&=F_2(y, z^S(x)) ,
\end{aligned}
\end{equation}
where $\mathcal{L}_{TSD}$ contains three losses for distillation ($L^{T_f}_{Lat}$), soft-label ($L^{T_f}_{Soft}$) and hard-label ($L_{Hard}$). $z^{T_f}$ and $z^S$ denotes the logit of finetuned teacher and student respectively. $y$ represents the ground-truth label from task-specific data. For supervised classification problems, we choose Kullback-Leibler Divergence \cite{Kullback51klDivergence} for $F_1$ and cross entropy for $F_2$. For regression task, we choose mean square error for both $F_1$ and $F_2$.

\subsection{Progressive Distillation Framework}
\vspace{-2mm}
The key technique for ERNIE-Tiny is to change the teacher, training data and learning objective carefully and smoothly. The whole distillation framework consists of general distillation, general-enhanced distillation, task-adaptive distillation and task-specific distillation. Overall, the student are trained using following four losses:
\begin{equation}
    \scalebox{0.9}{\begin{math}\begin{aligned}
        \mathcal{L}_{\{T, D, \alpha \}} = \mathop{\mathbb{E}}_{(x,y) \sim {D}} \mathcal{L}_{Lat}^{T}(x) + \alpha  (\mathcal{L}_{Soft}^{T}(x)+\mathcal{L}_{Hard}(x,y)) = \begin{cases}
\mathcal{L}_{GD}, &  T=T_g, D=D_g, {\alpha=0} \\
\mathcal{L}_{GED}, & T=T_f, D=D_g, {\alpha=0}  \\
\mathcal{L}_{TAD}, & T=T_f, D=D_t, {\alpha=0}  \\
\mathcal{L}_{TSD}, & T=T_f, D=D_t, {\alpha=1}  
\end{cases} 
\end{aligned}
\end{math}
   }
\end{equation}
where $T\in\{T_{g}, T_f\}$, $D\in \{D_g, D_t\}$ and $\alpha \in \{0,1\}$. Put them together, ERNIE-Tiny presents a smoothly transited distillation framework to effectively compress a large teacher model into a significantly smaller student model. As shown in Figure~\ref{fig:Distillation}, ERNIE-Tiny carefully designs these four stages such that only one of the three key components (i.e. training data, teacher and learning objective) changes between any two consecutive stages, while existing works (\cite{bert-emd, tinybert, distill-bert}) change all three of them. The advantage of each stage is shown in the ablation studies.
\vspace{-3mm}
\section{Experiment}
In this section, we first evaluate a 4-layer ERNIE-Tiny on English datasets and compare it with existing works. Then, we evaluate ERNIE-Tiny on Chinese datasets. After that, ablation studies and discussions are presented to analyze the contribution of each stage.
\subsection{Evaluation on English Datasets}
\subsubsection{Downstream Tasks} 
General Language Understanding Evaluation (GLUE) benchmark \cite{glue} is chosen to evaluate ERNIE-Tiny. It is a well-studied collection of NLP tasks, including textual entailment, emotion detection, coreference resolution, etc. 
In detail, we choose to evaluate our model in 8 tasks, namely Multi-Genre Language Inference(MNLI), Question Natual Language Inference(QNLI), Quora Question Paris (QQP),  Standford Sentiment Treebank(SST-2),  Semantic Textual Similarity Benchmark(STS-b), Corpus of Linguistic Acceptability(CoLA), Recognizing Textual Entailment(RTE), and Microsoft Research Paragraph Corpus(MRPC).

\subsubsection{Experiment Setup} \label{sec:experiment_setup}
For a fair comparison, we adopt pretrained BERT${}_{base}$ checkpoint released by the author~\cite{devlin2018bert} as pretrained teacher. BERT${}_{base}$ is a 12-layer transformer-based model with hidden size of 768 and intermediate size of 3072, accounting for 109M parameters in total, pretrained on English Wikipedia and BooksCorpus \cite{zhu2015aligning}. To obtain finetuned teachers, we finetune pretrained BERT${}_{base}$ on each task as finetuned teachers. Following existing works \cite{tinybert}, we adopt a 4-layer model with hidden size of 312 and intermediate hidden size of 1200 as our student. Besides GLUE as task-specific data, data (i.e. English Wikipedia and Bookscorpus) used to pretrain BERT${}_{base}$ are utilized as general data to ensure no additional resources or knowledge are involved. Recall that a finetuned teacher and general data are combined to perform distillation during GED. To adapt general data for these teacher finetuned on text pair task (i.e. MNLI, QQP,  QNLI, RTE, STS-B, MRPC), we construct input pairs by sampling consecutive text pairs in general data. 

\subsubsection{Results on English Datasets}
We compare ERNIE-Tiny with several baselines including BERT-PKD \cite{pkd}, DistilBERT \cite{distill-bert}, BERT-EMD \cite{bert-emd}, MobileBERT\cite{mobilebert} and TinyBERT \cite{tinybert}. The results of MobileBERT, TinyBERT and BERT-EMD are quoted from their paper. As BERT-PKD and DistilBERT do not experiment with 4-layer model, we quote the results from the TinyBERT's implementation \cite{tinybert}. We report test set results evaluated by the official GLUE server, summarized in Table~\ref{tab:glue}. Since MiniLM \cite{minilm} do not report test result on GLUE, we do not compare it. ERNIE-Tiny outperforms TinyBERT, DistilBERT, BERT-PKD, and BERT-EMD across most tasks and exceeds SOTA by 1.0\% GLUE score. Compared with its teacher BERT$_{base}$, a 4-layer ERNIE-Tiny retains 98.0\% performance while is 7.5x smaller and 9.4x faster for inference.
\vspace{-2mm}

\begin{table*}[htbp]
	\centering
	\scalebox{0.7}{
			\begin{tabular}{
					ccccccccccccc }
				\toprule
				Method         &Params  &Speedup          & MNLIm & MNLImm & QQP & SST-2 & QNLI & MRPC & RTE & CoLA & STS-B &  Avg. \\ 
				\midrule
				BERT${}_{Base}$(T.)   & 109M    &1x   & 84.6 & 83.4 & 71.2 & 93.5 & 90.5 & 88.9 & 66.4 & 52.1 & 85.8 & 79.6\\ 
				\midrule
				DistilBERT  & 52.2M         &3x        & 78.9 & 78.0 & 68.5 & 91.4 & 85.2 & 82.4 & 54.1 & 32.8 & 76.1 & 71.9 \\
			    BERT-PKD              & 52.2M    &3x    & 79.9 & 79.3 & 70.2 & 89.4 & 85.1 & 82.6 & 62.3 & 24.8 & 79.8 & 72.6 \\
				BERT-EMD 	 & 14.5M  &9.4x          & 82.1 & 80.6 & 69.3 & 91.0 & 87.2 & 87.6 & 66.2 & 25.6 & \textbf{82.3} & 74.7 \\
				MobileBERT* & 15.1M &8.6x &  81.5 & 81.6 & 68.9 & 91.7 & \textbf{89.5} & 87.9 & 65.1 &  46.7 & 80.1 &  77.0 \\ 
				TinyBERT          & 14.5M  &9.4x          & 82.5 & \textbf{81.8} & \textbf{71.3} & 92.6 & 87.7 & 86.4 & \textbf{66.6} & 44.1 & 80.4 & 77.0\\ 
				ERNIE-Tiny       & \textbf{14.5M}  &\textbf{9.4x}           & \textbf{83.0} & \textbf{81.8} & \textbf{71.3} & \textbf{93.3}    & 88.3 & \textbf{88.4} & \textbf{66.6} & \textbf{47.4} & \textbf{82.3} & \textbf{78.0} \\
				\bottomrule
			\end{tabular}
	}
	\caption{GLUE test results that are scored by GLUE evaluation server. The state-of-the-art results are in bold. All methods adopt BERT$_{base}$ as teacher model, excluding MobileBERT. MobileBERT* is distilled from IB-BERT, which has the same amount of parameters with BERT$_{large}$.
	The architecture of ERNIE-Tiny, BERT-EMD and TinyBERT is (${L}$=4, ${d}$=312, ${d}_{ff}$=1200), BERT-PKD and DistilBERT is (${L}$=4, ${d}$=768, ${d}_{ff}$=3072), MobileBERT is (${L}$=24, ${d}$=128, ${d}_{ff}$=512) with different transformer architecture design.
    	}
	\label{tab:glue}
\end{table*}
\vspace{-4mm}
    \subsection{Evaluation on Chinese Datasets}
    \paragraph{Dataset} 5 Chinese NLP datasets are chosen for evaluating ERNIE-Tiny, including XNLI \cite{conneau-etal-2018-xnli} for natural language inference, LCQMC \cite{liu-etal-2018-lcqmc} for semantic similarity, ChnSentiCorp\footnote{https://github.com/pengming617/bert\_classification} for sentiment analysis, NLPCC-DBQA\footnote{http://tcci.ccf.org.cn/conference/2016/dldoc/evagline2.pdf} for question answering and MSRA-NER \cite{zhang2006word-msraner} for named entity recognition. All results reported in this section are calculated by taking the average on the dev set result of 5 runs.
    \vspace{-5mm}
    \paragraph{Result} Since most of the compression models do not experiment on Chinese tasks, we reproduce TinyBERT for comparison. Both TinyBERT and ERNIE-Tiny are distilled from an strong teacher Chinese ERNIE2.0$_{base}$\cite{sun2019ernie20} instead of Chinese BERT$_{base}$. It can be seen in Table~\ref{tab:chinese} that ERNIE-Tiny outperforms Chinese BERT$_{base}$\footnote{https://github.com/google-research/bert} on XNLI, LCQMC, ChnSentiCorp and NLPCC-DBQA, and exceeds it by 0.4\% average score over the five datasets, while being 7.5x smaller and 9.4x faster in inference time. 
    
    	\begin{table*}[h]
		\begin{center}
			\scalebox{0.7}{
				\begin{tabular}{
						l|c|c|ccccc|c
					}
					\toprule
					Method   & Params  & Speedup     & XNLI   & LCQMC  & ChnSentiCorp     & NLPCC-DBQA  & MSRA-NER  & Avg.      \\
					\midrule
					ERNIE2.0${}_{Base}$ (T.) & 109M& 1x &79.8 & 87.5 & 95.5 & 84.4 & 95.0 & 88.4 \\
					\midrule
					BERT${}_{Base}$ & 109M & 1x &77.2 & 87.0 & 94.3 & 80.9 & \textbf{92.3} & 86.3   \\
				    TinyBERT (re.)& 14.5M & 9.4x & 76.3 & 86.8 & 94.2 & 81.8 & 87.3 &  85.3 \\
					ERNIE-Tiny &  \textbf{14.5M} & \textbf{9.4x} & \textbf{77.6} & \textbf{88.0} & \textbf{94.9} & \textbf{82.2} & 90.8 &  \textbf{86.7} \\
					\bottomrule
				\end{tabular}
			}
            \vspace{-1mm}
			\caption{Test Results of Chinese Tasks. TinyBERT on this table is reproducted by us. The teacher of TinyBERT and ERNIE-Tiny(${L}$=4, ${d}$=312, ${d}_{ff}$=1200) are set to Chinese ERNIE2.0$_{base}$. Both BERT${}_{base}$ and ERNIE2.0$_{base}$ is (${L}$=12, ${d}$=768, ${d}_{ff}$=3072).}
		    \label{tab:chinese}
		\end{center}
		\vspace{-5mm}
	\end{table*}

\subsection{Ablation Studies} \label{sec:ablation}
    We perform ablation studies on each stage involved in ERNIE-Tiny. To better illustrate the contribution of each stage, we divide them into two categories based on the training data used: GD and GED as general distillation; TAD and TSD as task-specific distillation. Experiments in this section follow experiment setup in Section~\ref{sec:experiment_setup}. All results in this section are obtained by taking the average on the dev set result of 5 runs.
	\vspace{-1mm}
	\paragraph{Effect of General Data Based Distillation}
	To analyze the contribution of general data, we perform ablation studies on 2 low-resource tasks MRPC/CoLA and 1 high-resource task MNLI. 
	As general data is utilized in GD and GED, we construct 4 different settings of ERNIE-Tiny to demonstrate the effect of distilling with general data by removing \textbf{1)} both GD and GED; \textbf{2)} GD; \textbf{3)} GED; \textbf{4)} None. 
	As shown in Table~\ref{tab:general_data_ablation}, removing both GD and GED significantly worsens the performance of distilled student, suggesting that general data plays an important role in distillation. 
	Recall that the only difference between GED and GD is that GED
	equips a finetuned teacher model. Compared with pretrained treacher, finetuned teacher captures task-specific information and is able to extract task-specific knowledge from general data. The results show that ERNIE-Tiny without GD exceeds the one without GED by 0.4\% average score, indicating that GED has a larger contribution than GD on distillation.
	\vspace{-1mm}

	\begin{table}[!ht]
	\begin{minipage}[t]{.48\linewidth}
		\scalebox{0.7}{
			\begin{tabular}[t]{ l|ccccc}
				\toprule
				Method          & MNLI(m/mm)                          & MRPC                    & CoLA        & Avg. \\
				\midrule
				BERT${\_}{Base}$ (T.)             & 84.5/84.6                & 86.8                  & 61.3  & 79.3    \\
				\midrule
				ERNIE-Tiny & 83.0/83.0                    & 86.9 & 50.0 & 75.7    \\
				\quad w/o GD   & 83.0/83.0  & 86.4           & 49.1   &75.4  \\
				\quad w/o GED    & 82.4/82.4                   & 86.1           & 49.2	& 75.0    \\

				\quad w/o GDD{\&}GED    & 63.9/63.6                   & 68.9          & 23.5	& 55.0   \\
				\bottomrule
			\end{tabular}
		}
		\vspace{1mm}
		\captionsetup{width=0.9\textwidth}
		\caption{Ablation study on distillation with general data on development set. (T.) denotes the teacher model.}
		\label{tab:general_data_ablation}
	\end{minipage}
	\begin{minipage}[t]{.48\linewidth}
		\scalebox{0.7}{
		\begin{tabular}[t]{ l|ccccc } 
			\toprule
			Method & MNLI(m/mm)                  & MRPC         &   CoLA & Avg.  \\
							\midrule

			BERT${\_}{Base}$ (T.)             & 84.5/84.6                & 86.8                  & 61.3  & 79.3    \\
			\midrule
			ERNIE-Tiny    &83.0 /83.0       &86.9 &  50.0 & 75.7 \\ 
			\quad w/o TAD        &81.2 /81.5       &     86.3   &  37.4 & 71.6\\ 
			\quad w/o TAD{\&}TSD + FT   &80.7 / 81.2         &79.8 &  $0^{\star}$ &60.4      \\ 
			\bottomrule
		\end{tabular}
		}
		\vspace{1mm}
		\caption{Ablation study on distillation with task-specific data. FT denotes finetuning model directly. The MCC of CoLA is 0$^{*}$  means no better than random prediction.}
		\label{tab:task_data_ablation}
	\end{minipage}
	\vspace{-3mm}

\end{table}

	\vspace{-4mm}
	\paragraph{Effect of Task-specific Data Based Distillation}
	To demonstrate the effectiveness of distillation on task-specific data, we vary the training process when performing distillation on task-specific data and summarize the results in Table~\ref{tab:task_data_ablation}. 
The results show that solely removing TAD consistently leads to a performance drop across all tasks. 
Note that TAD only differs from TSD in that TAD has only $\mathcal{L}_{Lat}$ involved while the loss in TSD comprises $\mathcal{L}_{Lat}$, 
$\mathcal{L}_{Soft}$ and $\mathcal{L}_{Hard}$. 
The results verify that the transition smoothing brought by TAD is crucial to the effectiveness of distillation. 
We then remove distillation on task-specific data entirely (i.e. TAD and TSD) and only finetune student of task-specific data, and find significant performance degradation. 
This indicates that distillation on task-specific data is non-negligible.

	\vspace{-2mm}
	\paragraph{Effect of Student Capacity}
	To illustrate the effect of the student model size, we enlarge the size of the student model to have the same size as the teacher model. As shown in Table~\ref{tab:self_train}, an ERNIE-Tiny with the original model size can exceed the teacher by 0.8\% average score. 
		\begin{table*}[htbp]
	\centering
	\scalebox{0.7}{
		\begin{tabular}{ l|ccccc|c }
			\toprule
			Method                        & MNLIm & MNLImm   & RTE   & MRPC  & CoLA  & Avg. \\ \midrule
			BERT${}_{Base}$ ($L$=12;$d$=768;${d}_{ff}$=3072) (T.)            & 84.5 & 84.6  & 69.7 & 86.8 & 61.3 & 77.4 \\
			\midrule 
			ERNIE-Tiny ($L$=4;$d$=312;${d}_{ff}$=1200)   & 83.0& 83.0  & 66.8 & 86.9 & 50.0 & 73.9 (-3.5)      \\ 
			ERNIE-Tiny ($L$=12;$d$=768;${d}_{ff}$=3072)  & 84.6&84.9  & 72.3 & 87.3 & 62.1 & 78.2 (+0.8)          \\ 
			\bottomrule
		\end{tabular}
		}
		\vspace{-1mm}
		\caption{Ablation study on student capacity. (T.) is the teacher model.}
		\label{tab:self_train}
	\end{table*}
	\vspace{-7mm}

	\subsection{Discussion}
	\vspace{-2mm}
	In this section, we analyze how general-enhanced distillation benefits the effectiveness of distillation. Experiments in this section follows the setup in Section~\ref{sec:experiment_setup}. All results are obtained by taking the average on the dev set results of 5 runs.
    \begin{table}[!htb]
\begin{minipage}{.4\linewidth}
		\centering
		\scalebox{0.7}{
		\begin{tabular}{ r|rrr }
			\toprule
			proportion  & MNLI  & QNLI  & QQP      \\
			\midrule
			1\%         & 3927  & 1047  & 3638      \\
			10\%        & 39270 & 10474 & 36384     \\
			50\%        & 196351& 52371 & 181925    \\
			\bottomrule
		\end{tabular}}
		\vspace{1mm}
	    \captionsetup{width=0.95\textwidth}
		\caption{Number of labeled data.}
		\label{tab:datasize}
		\vspace{4mm}
		\scalebox{0.7}{
		\begin{tabular}{l|c|cc}
			\toprule
			Method                                 &     MNLIm   & SNLI & RTE  \\
			\midrule 
			GD+GED+TSD      &    81.2     & 75.9  & 65.7 \\ 
			GD+TAD+TSD       &  82.4 & 70.9 & 52.8\\  
            GD+TSD       &   80.8 & 63.6 & 47.3\\  
			\bottomrule
		\end{tabular}}
		\vspace{1mm}
		\captionsetup{width=0.85\textwidth}
		\caption{Accuracy on out-of-domain datasets.}
		\label{tab:ood}
    \end{minipage}%
\begin{minipage}{.6\linewidth}
	\scalebox{0.7}{
		\begin{tabular}{ l|cccc|c }
			\toprule
			Method                           & MNLIm & MNLImm         & QNLI         & QQP    & Avg.  \\
			\midrule
			\multicolumn{4}{l}{1\% of labeled data}                                             \\
			\hline
			BERT${}_{base}$ (T.)           & 67.0 & 69.3       & 78.4       & 71.3  & 71.5\\
			ERNIE-Tiny(w/o GED)                   & 57.7& 60.5	 & 75.4       & 69.4 & 65.8  \\
			ERNIE-Tiny(w/ GED)                   & 65.2 & 67.4     & 75.4       & 70.8  & 69.7 \\
			gain of GED                          &  +7.5& +6.9       & +0.0       & +1.4 & +4.0  \\
			\midrule
			\multicolumn{4}{l}{10\% of labeled data}                                             \\
			\hline
			BERT${}_{base}$ (T.)            & 76.4&77.3	     & 86.9       & 79.7  & 80.1 \\
			ERNIE-Tiny(w/o GED)                   & 69.1&69.8      & 82.4       & 78.2 & 74.9  \\
			ERNIE-Tiny(w/ GED)                 & 74.5&75.0      & 82.4	    & 78.1 &77.5 \\
			gain of GED                           & +5.4&+5.2       & +0.0        & -0.1  & +2.6  \\
			\midrule
			\multicolumn{4}{l}{50\% of labeled data}                                             \\
			\hline
			BERT${}_{base}$ (T.)            & 80.5&81.9	     & 90.1       & 84.2  & 84.2 \\
			ERNIE-Tiny(w/o GED)                  & 75.3&76.4	     & 83.5       & 83.3 & 79.6  \\
			ERNIE-Tiny(w/ GED)                   & 79.3&80.1	     & 84.2       & 83.5  & 81.8 \\
			gain of GED                           & +4.0&+3.7       & +0.7       & +0.2   & +2.2  \\
			\bottomrule
		\end{tabular}
		}
		\vspace{1mm}
		\caption{Ablation study on labeled data size.}
		\label{tab:datasize_performance}
		\end{minipage}
		\vspace{-7mm}
\end{table}

	\paragraph{General Data as Supplement to Task-specific Data} ERNIE-Tiny transfers task-specific knowledge from finetuned teacher over \textit{general data} to student model in GED, while it transfers task-specific knowledge over \textit{task-specific data} in TAD and TSD. General data in GED can be regarded as a supplement to task-specific data. The effect of additional data should be more significant on low-resource tasks. To illustrate this, we vary the task-specific dataset size of MNLI, QNLI, and QQP tasks to 1\%, 10\%, and 50\% of the original size, the resulting data sizes are listed in Table \ref{tab:datasize}. Then we finetune BERT$_{base}$ to obtain finetuned teacher and perform distillation on student model with the finetuned teacher for each configuration. 
	Results are presented in Table \ref{tab:datasize_performance}, from which we can see that the gain from GED is more significant when less task-specific data is used.

    \vspace{-2mm}
    
    \paragraph{Generalization Gain by GED} Besides its benefits on low-resource tasks, GED can also be considered as a stage to improve the generalization of the student, as it allows the student to capture task-specific knowledge on a much larger dataset. Several works \cite{laine2016temporal, sajjadi2016regularization, miyato2018virtual, goodfellow2014explaining} succeeded in using random image augmentation to improve generalization performance for semi-supervised tasks. Similarly, at this stage, the hidden representation information still contains task-specific data distribution information, which can be used to compensate for the sparse task data and augment the feature representations. This leads to improving the generalization of the student model. To show that, we first distill ERNIE-Tiny on MNLI and then evaluate it on out-of-domain datasets including SNLI \cite{bowman-etal-2015-large-snli} and RTE. As RTE is a 2-class classification task while MNLI is a 3-class classification task, we simply drop the "neural" and take argmax of "entailment" and "not entailment" when calculating accuracy on RTE. As shown in Table~\ref{tab:ood}, experiment with GED exceed those without GED by a large margin. In particular, although removing one of GED or TAD results in similar MNLI accuracy, the experiment with GED significantly outperforms the one without GED on all out-of-domain datasets, demonstrating the generalization benefit led by GED. Another interesting observation is that adding TAD can also be beneficial to the generalization of the student.
    


\vspace{-2mm}
\section{Conclusion}
\vspace{-3mm}
In this paper, we propose a progressive distillation framework ERNIE-Tiny to compress PLMs. Four-stage distillation is introduced to smooth the transition from pretrain distillation to finetune distillation. 
In particular, general-enhanced distillation employs finetuned teacher to deliver enhanced knowledge over general data to student model, boosting the generalization of student model.
Task-adaptive distillation further smooths transition via carefully designed learning objectives.
ERNIE-Tiny distilled from BERT$_{base}$ retains 98\% performance with 9.4x faster inference speed, achieving SOTA on GLUE benchmark with the same amount of parameters.
Our 4-layer ERNIE-Tiny distilled from Chinese ERNIE2.0$_{base}$ also outperforms 12-layer Chinese BERT$_{base}$.
Our work didn't apply larger unlabeled general data such as C4\cite{t5}. More efficient data utilization is left for future work.
\vspace{-2mm}
\section{Broader Impact}
\vspace{-2mm}
Pretrained language models have recently achieved great success on a variety of natural language processing tasks. 
However, their enormous number of parameters lead to high storage and computational demands, bringing challenges to deploying and serving them on real-life applications. 
Our work targets on addressing this problem by compressing large pretrained model into a significantly smaller and faster student model while maintaining accuracy. On the one hand, Inferencing with compressed model can significantly reduce GPU hours required, energy consumption and carbon dioxide emissions. On the other hand, our method relaxes the requirement on high-end GPU for low-latency inference and allows them to run on low-resource devices. The relaxation also benefits researchers with fewer computational resources.

{\small
	\bibliographystyle{plain}
	\bibliography{reference}
}	

\appendix

\section{Appendix}

\subsection{Implementatoin Details}
Table\ref{tab:hyper} gives the detailed hyper-parameter used in our ablation experiment on GLUE tasks, the GED stage takes about 2 days on 4 16g-V100, while the computation cost for other tasks tasks no more than 1 day. Hyper-parameter for our chinese results is listed in Table\ref{tab:hyper-chn}.

	\begin{table*}[htbp]
	    \centering
		\begin{tabular}{ l|ccc }
			\toprule
			Hyper parameter                           & MNLI,QQP,QNLI & RTE,SST-2,STS-b & CoLA  \\
			\midrule
			\multicolumn{4}{l}{GED}                                             \\
			\hline
			batch size                        &  \multicolumn{3}{c}{1280} \\
			learning rate                     &  \multicolumn{3}{c}{2e-4} \\
			training steps                    &  \multicolumn{3}{c}{500K} \\
			optimizer                         &  \multicolumn{3}{c}{Adam} \\
			warmup steps                      &  \multicolumn{3}{c}{500} \\
            dropout rate                      &  \multicolumn{3}{c}{0.} \\
			\midrule
			\multicolumn{4}{l}{TAD}                                 \\
			\hline
			batch size                           & 256   & 128	    &128 \\
			learning rate                        & 5e-5  & 5e-5     &5e-5 \\
			epoch                                & 5    & 10       &50 \\
			optimizer                            & \multicolumn{3}{c}{Adam} \\
			warmup proportion                    &  \multicolumn{3}{c}{0.1} \\
            dropout rate                         &  \multicolumn{3}{c}{0.} \\
			\midrule
			\multicolumn{4}{l}{TSD}                                 \\
			\hline
			batch size                           & 256   & 128      &128 \\
			learning rate                        & 1e-5  & 3e-5     &3e-5 \\
			epoch                                & 3     & 3        &3 \\
			optimizer                            &  \multicolumn{3}{c}{Adam} \\
			warmup proportion                    &  \multicolumn{3}{c}{0.1} \\
            dropout rate                         &  \multicolumn{3}{c}{0.} \\
			\bottomrule
		\end{tabular}
		\caption{Hyper-parameter for ablation studies on GLUE tasks}
		\label{tab:hyper}
\end{table*}

\begin{table*}[htbp]
	    \centering
		\begin{tabular}{ l|ccccc }
			\toprule
			Hyper parameter       & XNLI &chnsenticorp&msra-ner&lcqmc&dbqa\\
			\midrule
			\multicolumn{4}{l}{GED}                                             \\
			\hline
			batch size                        &  \multicolumn{5}{c}{1280} \\
			learning rate                     &  \multicolumn{5}{c}{2e-4} \\
			training steps                    &  \multicolumn{5}{c}{500K} \\
			optimizer                         &  \multicolumn{5}{c}{Adam} \\
			warmup steps                      &  \multicolumn{5}{c}{500} \\
            dropout rate                      &  \multicolumn{5}{c}{0.} \\
			\midrule
			\multicolumn{4}{l}{TAD}                                 \\
			\hline
			batch size        & 256   & 128   & 256   & 128	    &128 \\
			learning rate     & 5e-5  & 5e-5   & 5e-5  & 5e-5     &5e-5 \\
			epoch             & 5    & 10    & 5    & 10       &50 \\
			optimizer         & \multicolumn{5}{c}{Adam} \\
			warmup proportion &  \multicolumn{5}{c}{0.1} \\
            dropout rate      &  \multicolumn{5}{c}{0.} \\
			\midrule
			\multicolumn{4}{l}{TSD}                                 \\
			\hline
			batch size       & 256   & 128       & 256   & 128      &128 \\
			learning rate    & 1e-5  & 3e-5        & 1e-5  & 3e-5     &3e-5 \\
			epoch            & 3     & 3         & 3     & 3        &3 \\
			optimizer          &  \multicolumn{5}{c}{Adam} \\
			warmup proportion  &  \multicolumn{5}{c}{0.1} \\
            dropout rate       &  \multicolumn{5}{c}{0.} \\
			\bottomrule
		\end{tabular}
		\caption{Hyper-parameter for chinese experiment}
		\label{tab:hyper-chn}
\end{table*}

\end{document}